\pdfoutput=1

\documentclass[11pt]{article}

\usepackage{emnlp2021}

\usepackage{times}
\usepackage{latexsym}
\usepackage[inline]{enumitem}
\usepackage{amsmath}
\usepackage{amssymb}
\usepackage{graphicx}
\usepackage{booktabs}
\usepackage{pifont}
\usepackage{footmisc}

\usepackage[T1]{fontenc}

\usepackage[utf8]{inputenc}

\usepackage{microtype}

\title{Open Knowledge Graphs Canonicalization using \\ Variational Autoencoders}

\author{Sarthak Dash, Gaetano Rossiello, Sugato Bagchi, \\ {\bf Nandana Mihindukulasooriya, Alfio Gliozzo} \\ 
    IBM Research AI, Thomas J. Watson Research Center \\ 
    Yorktown Heights, NY \\ 
    } 

\makeatletter

\begin{document}
\maketitle
\begin{abstract}
Noun phrases and Relation phrases in open knowledge graphs are not canonicalized, leading to an explosion of redundant and ambiguous subject-relation-object triples. Existing approaches to solve this problem take a two-step approach. First, they generate embedding representations for both noun and relation phrases, then a clustering algorithm is used to group them using the embeddings as features. In this work, we propose Canonicalizing Using Variational Autoencoders (CUVA)\footnote{\url{https://github.com/IBM/Open-KG-canonicalization}}, a joint model to learn both embeddings and cluster assignments in an end-to-end approach, which leads to a better vector representation for the noun and relation phrases. Our evaluation over multiple benchmarks shows that CUVA outperforms the existing state-of-the-art approaches. Moreover, we introduce \textsc{CanonicNell}, a novel dataset to evaluate entity canonicalization systems.
\end{abstract}

\section{Introduction}


Open Information Extraction (OpenIE) methods \cite{fader-etal-2011-identifying,stanovsky-etal-2018-supervised} can be used to extract triples in the form \emph{(noun phrase, relation phrase, noun phrase)} from given text corpora in an unsupervised way without requiring a pre-defined ontology schema.
This makes them suitable to build large Open Knowledge Graphs (OpenKGs) from huge collections of unstructured text documents, thereby making the usage of OpenIE methods highly adaptable to new domains. 

Although OpenIE methods are highly adaptable, one major shortcoming of OpenKGs is that Noun Phrases (NPs) and Relation Phrases (RPs) are not \emph{canonicalized}. This means that two NPs (or RPs) having different surface forms, but referring to the same entity (or relation) in a canonical KB, are treated differently. Consider the following triples as an example: (NBC-TV, has headquarters in, NYC), (NBC Television, is in, New York City) and (NBC-TV, has main office in, NYC). Looking at the previous example, both OpenIE methods and associated Open KGs would not have any knowledge that \emph{NYC} and \emph{New York City} refer to the same entity, or \emph{has headquarters in} and \emph{has main office in} are similar relations. 

Moreover, while it is true that similar relations will have same argument types (see the previous example), the converse need not hold true. For example, given the following two triples (X, is born in, Y) and (X, has died in, Y) in an Open KG, where X is of type \emph{Person} and Y is of type \emph{Location}, does not imply \emph{is born in} and \emph{has died in} are similar relations.

Thus, the task of \emph{canonicalizing} NPs and RPs within an Open KG is significant. Otherwise, Open KGs will have an explosion of redundant facts, which is highly undesirable, for the following reasons. Firstly, redundant facts use a higher memory footprint. Secondly, querying an Open KG is likely to yield sub-optimal results, for e.g. it will not return all facts associated with \emph{NYC} when using \emph{New York City} as the query. Finally, allowing downstream applications such as Link Prediction \cite{DBLP:conf/nips/BordesUGWY13} to know that \emph{NYC} and \emph{New York City} refers to the same entity, will improve their performance while operating on large Open KGs. Hence, it is imperative to \emph{canonicalize} NPs and RPs within an Open KG. 

In this paper, we introduce Canonicalizing Using Variational Autoencoders (CUVA), a neural network architecture that learns unique embeddings for NPs and RPs as well as cluster assignments in a joint fashion. CUVA combines \begin {enumerate*} [label=\itshape\alph*\upshape)] \item The Variational Deep Embedding (VaDE) framework \cite{DBLP:conf/ijcai/JiangZTTZ17}, a generative approach to Clustering, and \item A KG Embedding Model that aims to utilize the structural knowledge present within the Open KG \end {enumerate*}. In addition, CUVA uses additional contextual information obtained from the documents used to build the Open KG. 

The input to CUVA is \begin {enumerate*} [label=\itshape\alph*\upshape)] \item An Open KG expressed as a list of triples and \item Contextual Information obtained from the documents \end {enumerate*}. The output is a set of NP and RP clusters grouping all items together that refer to the same entity (or relation).

In summary, we make the following contributions,
\begin{itemize}
    \item We introduce CUVA, a novel neural architecture for the \textsc{Canonicalization} task, based on joint learning of mention representations and cluster assignments for entity and relation clusters using variational autoencoders.
    \item We demonstrate empirically that CUVA improves state of the art (SOTA) on the Entity \textsc{Canonicalization} task, across four academic benchmarks.
\end{itemize}

\section{Related Work}
\label{relatedWork}
Extracting triples from sentences is the first step to build Open KGs. The OpenIE technique has been originally introduced in \cite{DBLP:conf/ijcai/BankoCSBE07}. Thereafter, several approaches have been proposed to improve the quality of the extracted triples. Rule-based approaches, such as \textsc{ReVerb}~\cite{fader-etal-2011-identifying} and \textsc{PredPatt}~\cite{white-etal-2016-universal}, use patterns on top of syntactic features to extract relation phrases and their arguments from text. Learning-based methods, such as \textsc{OLLIE}~\cite{mausam-etal-2012-open} and \textsc{RnnOIE}~\cite{stanovsky-etal-2018-supervised}, train a self-supervised system using bootstrapping techniques. Clause-based approaches~\cite{angeli-etal-2015-leveraging} navigate through the dependency trees to split the sentences into simpler and independent segments.

There have been several previous works to group NPs and RPs into coherent clusters. A traditional approach to canonicalize NPs is to map them to an existing KB such as Wikidata, also referred to as the Entity Linking (EL) task \cite{DBLP:conf/naacl/LinME12,DBLP:conf/semweb/CeccarelliLOPT14}. A major problem with these EL approaches is that many NPs may refer to entities that are not present in the KB, in which case they are not clustered. 

The RESOLVER system \cite{DBLP:journals/jair/YatesE09} uses string similarity features to cluster phrases in TextRunner \cite{DBLP:conf/ijcai/BankoCSBE07} triples. \cite{DBLP:conf/cikm/GalarragaHMS14} uses manually defined features for NP canonicalization, and subsequently performs relation phrase clustering by using AMIE algorithm \cite{DBLP:conf/www/GalarragaTHS13}. \cite{DBLP:conf/cikm/WuWKY18} propose a modification to the previous approach by using pruning and bounding techniques. Concept Resolver \cite{krishnamurthy-mitchell-2011-noun}, which makes ``one sense per category'' assumption, is used for clustering NP mentions in NELL. This approach requires additional information in the form of a schema of relation types.
KB-Unify \cite{delli-bovi-etal-2015-knowledge} addresses the problem of unifying multiple canonical and open KGs into one KG, but requires additional sense inventory, which may not be available.

The CESI architecture \cite{DBLP:conf/www/VashishthJT18} models the \textsc{Canonicalization} task in a two-step pipeline approach, i.e., in the \emph{first} step, it uses a HolE algorithm \cite{DBLP:conf/aaai/NickelRP16} to learn embeddings for NPs (and RPs), and then in an \emph{independent} second step, it ``plugs" these learned embeddings into a Hierarchical Agglomerative Clustering (HAC) algorithm to generate clusters. Currently, CESI is the state of the art on this task. Unlike CESI, our proposed model CUVA learns the embedding representations and the cluster assignments of both NPs and RPs in an end-to-end manner, using a single model.


\section{Open KGs Canonicalization Using VAE} \label{approach}

\begin{figure*}[t]
  \centering
  \includegraphics[width=\textwidth]{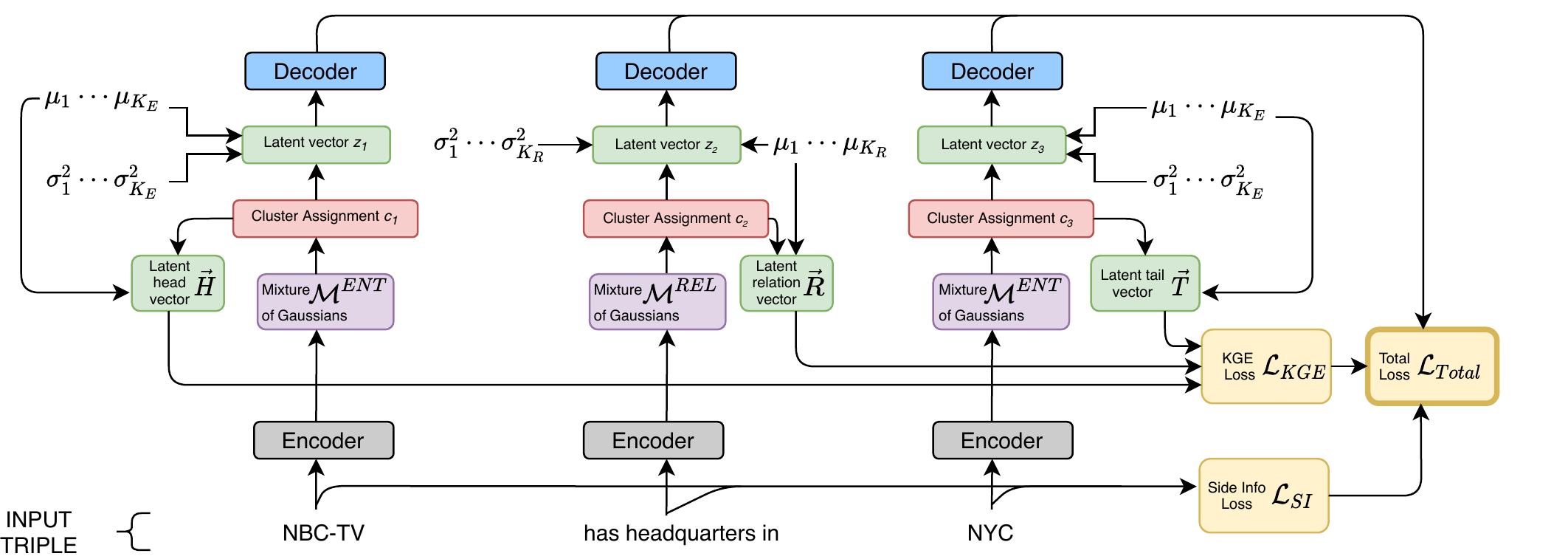}
  \caption{The core structure of CUVA. The \emph{left} and \emph{right} vertical structures correspond to the Entity variational autoencoder (E-VAE) for the head and tail Noun Phrases, whereas the \emph{middle} vertical structure corresponds to the Relation variational autoencoder (R-VAE) for the Relation Phrase. The KGE module connects both the E-VAE and the R-VAE as shown above.} 
  \label{architecture}
\end{figure*}

Formally, the \textsc{Canonicalization} task is defined as follows: given a list of triples $\mathcal{T} = (h,r,t)$ from an OpenIE system $\mathcal{O}$ on a document collection $\mathcal{C}$, where $h,t$ are Noun Phrases (NPs) and $r$ is a Relation Phrase (RP), the objective is to cluster NPs (and RPs), so that items referring to the same entity (or relation) are in the same cluster. 


We assume that each cluster corresponds to either a \emph{latent} entity or a \emph{latent} relation; the label of such a \emph{latent} entity/relation is unknown to the learner.

CUVA uses two variational autoencoders i.e. E-VAE and R-VAE, one each for \emph{entities} and \emph{relations}. Both E-VAE and R-VAE use a mixture of Gaussians for modeling \emph{latent} entities and relations. 
Also, we use a Knowledge Graph Embedding (KGE) module to encode the structural information present within the Open KG. CUVA works as follows: 
\begin{enumerate}
    \item A \emph{latent} entity (or relation) as defined above, is modeled via a Gaussian distribution. The sampled items from the Gaussian distribution correspond to the observed NPs (and RPs) within $\mathcal{T}$.
    \item NPs ($h,t$) and RPs ($r$) are modeled using larger embedding dimensions compared to the Gaussian distributions, to account for variations in the observed surface forms. 
    \item We use Gaussian parameters to refer to the entity (relation) as opposed to the NP (or RP).
    \item Since the items are clustered together, we assume that different NPs, e.g. \emph{New York City} and \emph{NYC} (or RPs) belonging to the same Gaussian distribution (i.e. cluster) have similar \emph{attributes}. 
\end{enumerate} 

Fig. \ref{architecture} illustrates an instantiation for CUVA. A description of each of the components of CUVA follows below.

\subsection{Variational Autoencoder} \label{vae_theory}

Based on the above modeling assumptions, we use Variational Deep Embedding (VaDE) \cite{DBLP:conf/ijcai/JiangZTTZ17} generative model for clustering. This generative clustering model implements a Mixture of Gaussians within the latent space of a variational autoencoder (VAE). We believe such a model is better suited to cluster mentions because its soft-clustering ability can account for different senses (polysemy) of a given entity mention. Such behavior is preferable to hard-clustering methods, such as agglomerative clustering algorithms, that assign each entity mention to exactly one cluster. Moreover, the high dimensional input space of VAE is better equipped to encode variations in the observed surface forms of different entity/relation mentions. 

The generative process of VaDE is described as follows. Assuming that there are $K$ clusters, an observed instance $x \in \mathbb{R}^D$ is generated as,
\begin{enumerate} 
\item Choose a cluster $c \thicksim \text{Cat}(\pi)$, i.e.\ a categorical distribution parametrized by probability vector $\pi$.
\item Choose a latent vector $z \thicksim \mathcal{N}(\mu_c, \sigma^2_c\textbf{I})$ i.e. sample $z$ from a multi-variate Gaussian distribution parametrized by mean $\mu_c$ and diagonal covariance $\sigma^2_c\textbf{I}$.
\item Compute $[\mu_x; \log \sigma^2_x] = f_\theta(z)$ where $f_\theta$ corresponds to a neural network parametrized by $\theta$, and $z$ is obtained from the previous step.
\item Finally, choose a sample $x \thicksim \mathcal{N}(\mu_x, \sigma^2_x\textbf{I})$ i.e.\ sample $x$ from a multi-variate Gaussian distribution parametrized by mean $\mu_x$ and diagonal covariance $\sigma^2_x\textbf{I}$.
\end{enumerate}

where $\pi_k$ is the prior probability for cluster $k$, $\pi \in \mathbb{R}^K_+$ and $\sum_{k=1}^K \pi_k = 1$. We make the same assumptions as made by \cite{DBLP:conf/ijcai/JiangZTTZ17}, and assume the variational posterior $q(z,c|x)$ to be a mean field distribution, and factorize it as:
\begin{equation}
    q(z,c|x) = q(z|x)q(c|x)
\end{equation}

We describe below the \emph{inner workings} of CUVA with respect to the \emph{head} Noun Phrase, or the leftmost vertical structure in Fig. \ref{architecture}. An analogous description follows for the \emph{tail} Noun Phrase, and the Relation Phrase as well. 

\begin{figure}[ht]
  \centering
  \includegraphics[width=\columnwidth]{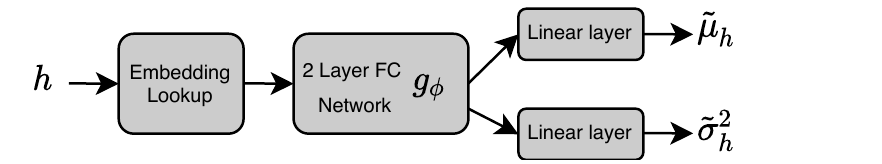}
  \caption{An Encoder block. CUVA uses separate fully connected networks $g^{E}_\phi$ and $g^{R}_\phi$ for E-VAE and R-VAE.} 
  \label{fig:encoder}
\end{figure}

\paragraph{Encoder:} Fig. \ref{fig:encoder} illustrates the Encoder block graphically. A Noun Phrase $h$, is fed as input to the Encoder block, which consists of: \begin {enumerate*} [label=\itshape\alph*\upshape)] \item An embedding lookup table, \item A \emph{two} layer fully connected neural network $g^{E}_\phi$ ($g^{R}_\phi$ for R-VAE) with \emph{tanh} non-linearity, and \item Two linear layers in parallel. \end {enumerate*} .

The Encoder block is used to model $q(z|h)$ i.e.\ the variational posterior probability of the latent representation $z$ given input representation $h$, via the following equations,
\begin{align}
    [\tilde{\mu}_h; \log \tilde{\sigma}^2_h] &= g^{E}_\phi(h)\\ 
    q(z|h) &= \mathcal{N}(z;\tilde{\mu}_h,\tilde{\sigma}^2_h\textbf{I}) 
\end{align}

After the parameters $\tilde{\mu}_h,\tilde{\sigma}_h$ for the variational posterior $q(z|h)$ have been calculated, we use the reparametrization trick \cite{DBLP:journals/corr/KingmaW13} to sample $z_1$ as follows, 
\begin{equation} \label{reparametrization}
    z_1 = \tilde{\mu}_h + \tilde{\sigma}_h \circ \epsilon 
\end{equation}
where $\epsilon \thicksim \mathcal{N}(0, \textbf{I})$ (i.e. a standard normal distribution) and $\circ$ denotes element-wise multiplication.

\begin{figure}[ht]
  \centering
  \includegraphics[width=\columnwidth]{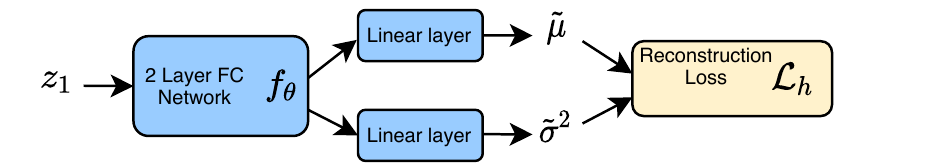}
  \caption{An Decoder block. CUVA uses separate fully connected networks $f^{E}_\theta$ and $f^{R}_\theta$ for E-VAE and R-VAE.} 
  \label{fig:decoder}
\end{figure}

\paragraph{Decoder:} Given $z_1$, the \emph{decoding} phase continues through the Decoder block, as illustrated in Fig. \ref{fig:decoder}, and via the following equations,
\begin{align}
    [\tilde{\mu}; \log \tilde{\sigma}^2] &= f^{E}_\theta(z_1)\\ 
    h^{\prime} &\thicksim \mathcal{N}(\tilde{\mu},\tilde{\sigma}^2\textbf{I})
\end{align}

Following \cite{DBLP:conf/ijcai/JiangZTTZ17}, the variational posterior $q(c|h)$ i.e. the probability of the NP $h$ belonging to cluster $c$ is calculated as: 
\begin{equation} \label{clust_assignment}
    q(c|h) = \frac{p(c)p(z|c)}{\sum_{c^\prime=1}^K p(c^\prime)p(z|c^\prime)}
\end{equation}
In practice, we use $z_1$ obtained from Equation \ref{reparametrization} in place of $z$ (in Equation \ref{clust_assignment}), and calculate a vector of assignment probability for an input $h$. 

During \emph{inference} phase, $h$ is assigned to a cluster having the highest probability, i.e.\ cluster assignment (in Fig. \ref{architecture}) occurs via a \emph{winners-take-all} strategy.

\subsection{The KGE Module} \label{kbc}
The \emph{motivation} behind using a Knowledge Graph Embedding (KGE) module is to encode the structural information present within the Open KG. This module is responsible for the joint learning between the latent representations for entities and relations (See Figure \ref{architecture}) and is described as follows. 


Given a triple mention $(h,r,t)$ belonging to an Open KG, we use Equation \ref{clust_assignment} to obtain a vector of cluster assignment probabilities $c^h, c^r$ and $c^t$ for the NPs and RPs respectively. As the next step, we choose a base $\tau > 0$ and employ a soft argmax function on probability vectors $c^h, c^r$ and $c^t$ as follows, 
\begin{align}
\label{soft-argmax}
    \sigma(c_\alpha) &= \frac{e^{\tau c^\alpha_i}}{\sum_{j=1}^K e^{\tau c^\alpha_j}} \text{, for } i = 1, \dots, K
\end{align}
where $\alpha \in \{h,r,t\}$ and $K$ denotes the number of clusters. 

Choosing a large value of $\tau$ ensures that the resulting vectors $v_h, v_t$ and $v_r$ obtained from Equation \ref{soft-argmax} are one-hot in nature, and indicate the most probable cluster ids for the NPs $h, t$ and RP $r$ respectively. For all our experiments, we choose $\tau=1e5$. In short, Equation \ref{soft-argmax} is a differentiable approximation to the non-differentiable argmax function. 

Given that, we now know the most probable cluster ids for a triple mention $(h,r,t)$, we build the \emph{entity} and \emph{relation} representations of these mentions, namely $e_h, e_t$ and $e_r$ as,
\begin{equation}
    e_h = v_h M_{E} \quad e_t = v_t M_{E} \quad e_r = v_r M_{R}
\end{equation}

where $M_{E}, M_{R}$ represent matrices containing \emph{mean} vectors (stacked across rows) for each of the $K_{E}$ and $K_{R}$ Gaussians present in E-VAE and R-VAE respectively. Here, $K_{E}$ and $K_{R}$ (Fig. \ref{architecture}) are hyper-parameters for CUVA. 

Once, we have the \emph{entity} and \emph{relation} representations, we use HolE as described in \citet{DBLP:conf/aaai/NickelRP16} as our choice of KGE algorithm for CUVA. 

\subsection{Side Information}\label{side_info_theory}

Noun and Relation Phrases present within an Open KG can be often tagged with relevant side information extracted from the context sentence in which the triple appears. We use the same side information (i.e. a list of equivalent mention pairs) as CESI \cite{DBLP:conf/www/VashishthJT18}. These side information tuples are obtained via the following sources/strategies: Entity Linking, PPDB (ParaPhrase DataBase), IDF Token Overlap, and Morph Normalization. Each source generates a list of equivalent mention pairs along with a score per pair. A description of these sources together with their associated scoring procedures is provided in Section \ref{appendix:SI} of the Appendix. 

\begin{figure}
  \centering
  \includegraphics[width=\columnwidth]{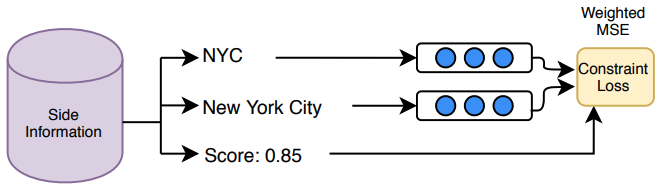}
  \caption{Encoding Side Information as a \emph{loss} measure.}
  \label{SIarchitecture}
\end{figure}

Let's consider the example of an \emph{equivalent} mention pair \emph{(NYC, New York City)} as shown in Fig. \ref{SIarchitecture} to illustrate the use of side information as a constraint in CUVA. We first perform an embedding lookup for the mentions \emph{NYC} and \emph{New York City} and then compute a Mean Squared Error (MSE) value weighted by its plausibility score. The MSE value indicates how far CUVA is from satisfying all the constraints represented as equivalent mention pairs. Finally, we sum up the \emph{weighted} MSE values for all equivalent mention pairs, which comprises our Side Information Loss $\mathcal{L}_{\text{SI}}$ in Fig. \ref{architecture}.

\section{Evaluation} \label{evaluation}

The \textsc{Canonicalization} task is \emph{inherently} unsupervised, i.e. we are not given any manually annotated data for training. With this in mind, we train the CUVA model according to the procedure described in Section \ref{appendix:trainingDetails} of the Appendix and then evaluate our approach on the \emph{Entity Canonicalization} task only. We do not include quantitative evaluations on the \emph{Relation Canonicalization} task, as none of the benchmarks described below have ground-truth annotations for canonicalizing relations, leaving the creation of a dataset for relation clustering as an interesting future work. 

\subsection{Benchmarks}

For comparing the performance of CUVA against the existing state of the art approaches, we use the Base and Ambiguous datasets introduced by \cite{DBLP:conf/cikm/GalarragaHMS14}, and ReVerb45K dataset from \cite{DBLP:conf/www/VashishthJT18}.

In addition, we introduce a \emph{new} dataset called \textsc{CanonicNell}, which we built by using the 165th iteration snapshot of NELL, i.e.\ Never-Ending Language Learner \cite{DBLP:conf/aaai/CarlsonBKSHM10} system. We created \textsc{CanonicNell} to build a dataset whose provenance is not related to ReVerb Open KB, unlike the datasets mentioned above. 

\paragraph{Building \textsc{CanonicNell}.} The \textsc{CanonicNell} dataset is built via an automated strategy as follows. The above snapshot of NELL aka NELL165, contains accumulated knowledge as a list of (subject, relation, object) triples. For building \textsc{CanonicNell}, we use the data artifact generated by \cite{DBLP:conf/semweb/PujaraMGC13} which marks co-referent entities within NELL165 triples, together with a soft-truth value per entity pair. We filter out all pairs having a score \emph{less} than $0.25$, and view the remaining pairs as undirected edges in a graph. To this graph, we apply a depth-first-search to obtain a set of connected components, which we refer to as the set of Gold Clusters. Next, we filter through the list of NELL165 triples and keep only those whose either \emph{head} or \emph{tail} entity is present within the set of \emph{Gold Clusters}. These triples together with the Gold Clusters obtained previously, form our newly proposed \textsc{CanonicNell} dataset. 


\begin{table}
    \centering
    \resizebox{.99\columnwidth}{!}{
    \begin{tabular}{l|p{0.19\columnwidth}|c|c|c}
    \toprule
    Datasets & Gold NP Clusters & NPs & RPs & Triples \\ \midrule
    Base & 150 & 290 & 3K & 9K \\ 
    Ambiguous & 446 & 717 & 11K & 37K \\ 
    ReVerb45K & 7.5K & 15.5K & 22K & 45K \\ 
    \textsc{CanonicNell} & 1.4K & 8.7K & 139 & 20K \\ 
    \bottomrule
    \end{tabular}}
    \caption{Details of datasets used. \textsc{CanonicNell} is the new dataset that we introduce in this paper.}
    \label{tab:Datasets}
\end{table}

\begin{table*}
    \centering
    \resizebox{.999\linewidth}{!}{
    \begin{tabular}{l|c|c|c|c|c|c|c|c|c|c}
    \toprule 
     & & \multicolumn{3}{c|}{Base Dataset} &  \multicolumn{3}{c|}{Ambiguous Dataset} &  \multicolumn{3}{c}{ReVerb45K} \\ \hline
    & SI & Macro & Micro & Pair & Macro & Micro & Pair & Macro & Micro & Pair \\ \midrule 
    $\text{Gal\'arraga-IDF}^{\dagger}$ & No & 0.948 & 0.979 & 0.983 & \textbf{0.679} & 0.829 & 0.793 & \textbf{0.716} & 0.508 & 0.005 \\
    $\text{GloVe+HAC}^{\dagger}$ & No & 0.957 & 0.972 & 0.911 & 0.659 & 0.899 & 0.901  & 0.565 & 0.829 & 0.753 \\ 
    $\text{GloVe+HAC+SI}$ & Yes & 0.972 & 0.998 & 0.999 & 0.665 & 0.898 & 0.764 & 0.666 & 0.847 & 0.708 \\ 
    $\text{HolE (GloVe)}^{\dagger}$ & No & 0.752 & 0.936 & 0.893 & 0.539 & 0.854 & 0.767 & 0.335 & 0.758 & 0.510 \\
    $\text{CESI}^{\dagger}$ & Yes & \textbf{0.982} & 0.998 & \textbf{0.999} & 0.662 & \bf{0.924} & 0.919 & 0.627 & 0.844 & 0.819 \\ \midrule 
    CUVA & Yes & \textbf{0.982} & \textbf{0.999} & \textbf{0.999} & 0.674 & \textbf{0.924} & \textbf{0.92} & 0.661 & \textbf{0.867} & \textbf{0.855} \\
    \bottomrule
    \end{tabular}}
    \caption{Macro, Micro and Pair F1 results on the \emph{Entity Canonicalization} task for \emph{head entity mentions}. Rows marked with a $\dagger$ are from \protect\cite{DBLP:conf/www/VashishthJT18}. CESI, the existing state-of-the-art approach is identical to HolE (GloVe) equipped with the Side Information. SI indicates whether an approach uses Side Information or not. Refer to Sec \ref{appendix:Hyperparameters} of the Appendix for the hyperparameter descriptions.}
    \label{ReVerb45kHeadEntities}
\end{table*}

Table \ref{tab:Datasets} shows the dataset statistics for all the benchmarks, wherein the split into test and validation folds for Base, Ambiguous and ReVerb45K datasets is already given by \citet{DBLP:conf/www/VashishthJT18}\footnote{\url{https://github.com/mallabiisc/cesi/}\label{cesi_url}}. This task is \emph{unsupervised} in nature, hence we do not possess any training data. For \textsc{CanonicNell}, we did a random 80:20 split of the triples into validation and test folds. For all methods, grid search over the hyper-parameter space using the validation set is performed, and results corresponding to the best-performing settings are reported on the test set. Following \citet{DBLP:conf/cikm/GalarragaHMS14}, we use the macro, micro, and pair F1 scores for evaluations.


Following \citet{DBLP:conf/www/VashishthJT18}, we use the Side Information as mentioned in Section \ref{side_info_theory} for canonicalizing NPs and RPs for the Base, Ambiguous, and ReVerb45K datasets. For canonicalizing NPs on \textsc{CanonicNell}, we use IDF Token Overlap as the only strategy to generate Side Information. This strategy is an inherent property of the dataset and needs no external resources (Section \ref{appendix:SI}). Moreover, for \textsc{CanonicNell} we do not canonicalize the RPs, since they are already unique.

Finally, a detailed description of the range of values tried per hyperparameter, and the final values used within CUVA is provided in Section \ref{appendix:Hyperparameters}.

\subsection{Results} \label{ent_results_reverb}

The existing state of the art model CESI \cite{DBLP:conf/www/VashishthJT18} evaluates on the Entity Canonicalization task using \emph{head entity mentions} only\footref{cesi_url}. To be comparable, we first evaluate on the head entity mentions only and illustrate our results in Table \ref{ReVerb45kHeadEntities}. Table \ref{ReVerb45kAllEntities} illustrates the results when evaluated for all entity mentions on Reverb45K. 

\begin{table}
    \centering 
    \begin{tabular}{l|c|c|c|c|c}
    \toprule 
     & \small SI & \small Macro & \small Micro & \small Pair & \small Mean \\ \midrule 
    CESI & Yes &  0.749 & 0.828 & 0.763 & 0.780 \\ 
    CUVA & Yes & \bf{0.755} & \bf{0.845} & \bf{0.81} & \bf{0.803} \\
    \bottomrule
    \end{tabular}
    \caption{Macro, Micro and Pair F1 results on the \emph{Entity Canonicalization} task for \emph{all entity mentions} on ReVerb45K. Both CESI and CUVA use the same hyperparameter settings as in Table \ref{ReVerb45kHeadEntities}.} 
    \label{ReVerb45kAllEntities}
\end{table}

The \emph{first} line in Table \ref{ReVerb45kHeadEntities}, i.e.\ Gal\'arraga-IDF \cite{DBLP:conf/cikm/GalarragaHMS14} depicts the performance of a feature-based method on this task. This approach is more likely to put two NPs together if they share a token with a high IDF value. The \emph{second} row in Table \ref{ReVerb45kHeadEntities}, i.e., GloVe+HAC uses a pretrained GloVe model \cite{pennington-etal-2014-glove-1} to first build embeddings for entity mentions and then uses a HAC algorithm for clustering. For multi token phrases, GloVe embeddings for tokens are averaged together. GloVe captures the semantics of NPs and does not rely on its surface form, thus performing well across all the datasets. The \emph{third} row augments GloVe+HAC by first initializing with pretrained GloVe vectors, followed by an optimization step wherein the Side Information (Section \ref{side_info_theory}) loss objective is minimized, and finally clustering via the HAC algorithm. The \emph{fourth} row, i.e. HolE (GloVe) uses the HolE Knowledge Graph Embedding model (initialized with pretrained GloVe vectors) to learn unique embeddings for NPs and RPs, followed by clustering using a HAC algorithm. It captures structural information with the KG and is an effective approach for NP Canonicalization.

The current state of the art, i.e.\ CESI \cite{DBLP:conf/www/VashishthJT18} extends the HolE (GloVe) approach, by adding Side Information (Section \ref{side_info_theory}) as an additional loss objective to be minimized. Looking at the results, it is clear that the addition of Side Information provides a \emph{significant} boost in performance for this task.


The final row in Table \ref{ReVerb45kHeadEntities} illustrates CUVA's performance on this task, which is an original contribution of our work. We observe that CUVA outperforms CESI on ReVerb45K and achieves the new state-of-the-art (SOTA). The improvement in the Mean F1 value, i.e., average over Macro, Micro, and Pair F1, over CESI is statistically significant with \emph{p value} being less than 1e-3. 

On the Ambiguous dataset, CUVA achieves a $1.2\%$ improvement over CESI on the Macro F1 metric, with the Micro and Pair F1 metrics achieving identical performance as CESI. Finally, on the Base dataset, CUVA achieves identical performance as CESI. Moreover, the results in Table \ref{ReVerb45kAllEntities} also show a similar trend when evaluated on all entity mentions, i.e., both head and tail NPs belonging to Reverb45K.

Table \ref{NellEntities} shows the results for the Entity Canonicalization task when evaluated on the \textsc{CanonicNell} dataset. The first two rows correspond to approaches that use pretrained FastText \cite{DBLP:conf/lrec/MikolovGBPJ18} and GloVe models to build unique embeddings for NPs and then use HAC to generate clusters. Moreover, in the absence of contextual information for the \textsc{CanonicNell} triples, both CESI and CUVA use IDF Token Overlap as the only source of Side Information. Moreover, from Table \ref{NellEntities}, it is clear that CUVA achieves the new state of the art result on this benchmark as well. 


\begin{table}
    \centering 
    \resizebox{.999\columnwidth}{!}{
    \begin{tabular}{l|c|c|c|c}
    \toprule 
     & SI & Macro & Micro & Pair \\ \midrule 
    FastText+HAC & No & 0.725 & 0.798 & 0.219 \\ 
    GloVe+HAC & No & 0.747 & 0.811 & 0.280 \\ 
    CESI & Yes & 0.749 & 0.817 & 0.307 \\ \midrule 
    CUVA & Yes & \bf{0.775} & \bf{0.826} & \bf{0.363} \\
    \bottomrule
    \end{tabular}}
    \caption{Macro, Micro and Pair F1 results on the \emph{Entity Canonicalization} task for the \textsc{CanonicNell} dataset. Refer to Section \ref{appendix:Hyperparameters} of the Appendix for the hyperparameters used in these experiments. SI indicates whether an approach uses Side Information or not.}
    \label{NellEntities}
\end{table}

\section{Qualitative Analysis} \label{QualitativeAnalysis}

Table \ref{tab:Predicted_Clusters} illustrates the output of our system for canonicalizing NPs and RPs on ReVerb45K. The top block corresponds to \emph{six} NP clusters, one per line. The algorithm is able to correctly group \emph{kodagu} and \emph{coorg} (different name of the same district in India), despite having completely different surface forms. However, a common mistake that our proposed system makes is depicted in row \emph{five}, i.e., four different people each having the same name \emph{bill} are clustered together. This error can be mitigated by keeping track of the \emph{type} information \cite{DBLP:conf/aaai/DashCGMF20} of each NP for disambiguation.

The bottom \emph{four} rows in Table \ref{tab:Predicted_Clusters} correspond to \emph{four} RP clusters. While the equivalence of RPs is captured in the first two rows of the bottom block, the \emph{final} two rows highlight a potential issue involving negations and antonyms, i.e. \emph{rank below} and \emph{be rank above} have opposite meanings. We leave the resolution of this issue as future work.

\begin{table}
    \centering
    \resizebox{.999\columnwidth}{!}{
    \begin{tabular}{l}
    \toprule 
    Predicted Clusters for ReVerb45K \\ \midrule 
     \{utc, coordinate universal time, universal coordinate time\} \ding{52} \\
     \{justice alito, samuel alito, sam alito, alito\} \ding{52} \\ 
     \{kodagu, coorg\} \ding{52} \\ 
     \{johnny storm, human torch\} \ding{52} \\ 
     \{bill cosby, bill maher, bill doolin, bill nye\} \ding{54} \\
     \{toyota, honda, toyota motor corporation\} \ding{54} \\ \midrule 
     \{be associate with, have be affiliate to, be now associate with\} \ding{52} \\ 
     \{lead a march on, lead the assault on\} \ding{52} \\ 
     \{be far behind, be not far behind, be way behind, \\ \qquad be close behind, be seat behind, be firmly entrench in \} \ding{54} \\ 
     \{rank below, rank just below, be rank above\} \ding{54} \\ 
    \bottomrule
    \end{tabular}}
    \caption{Examples of NP(\emph{top}) and RP(\emph{bottom}) clusters predicted by CUVA on ReVerb45K.}
    \label{tab:Predicted_Clusters}
\end{table}

\section{Further Analysis} 

In this section, we analyze CUVA under three different configurations. Section \ref{compare_with_bert} compares how CUVA performs against pretrained language models. Section \ref{structural_ablation} analyzes the effect of ablating components from our proposed network architecture. Finally, Section \ref{effectiveness_joint_learning} demonstrates the effectiveness of a joint learning approach over a pipeline-based strategy using the same network architecture.

\subsection{Comparison with Pretrained LMs} \label{compare_with_bert}

In this section, we investigate how CUVA fares against pretrained language models. Table \ref{tab:ReVerb45K_Bert} illustrates the results when evaluated on ReVerb45K. 

\begin{table}
    \centering 
    \resizebox{.999\columnwidth}{!}{
    \begin{tabular}{l|c|c|c|c}
    \toprule 
     & Macro & Micro & Pair & Mean \\ \midrule 
    RoBERTa+HAC & 0.448 & 0.804 & 0.776 & 0.676 \\
    BERT+HAC & 0.586 & 0.839 & 0.822 & 0.749 \\
    ERNIE+HAC & 0.591 & 0.842 & 0.825 & 0.753 \\ \midrule
    CUVA & \textbf{0.661} & \textbf{0.867} & \textbf{0.855} & \textbf{0.794} \\
    \bottomrule
    \end{tabular}}
    \caption{Macro, Micro and Pair F1 results for comparison with pretrained language models on the \emph{Entity Canonicalization} task for \emph{head entity mentions} on ReVerb45K.}
    \label{tab:ReVerb45K_Bert}
\end{table}

Following the observations of \cite{liu-etal-2019-linguistic,tenney-etal-2019-bert}, we use the lower layers (layers \emph{one} through \emph{six}) of a pretrained BERT, RoBERTa and a knowledge graph enhanced ERNIE \cite{zhang-etal-2019-ernie} base model via the HuggingFace Transformers library \cite{wolf-etal-2020-transformers}. For building static representation for each entity mentions, we use a mean pooling strategy to aggregate the contextualized representations. The entity mention representations are finally clustered using HAC. 

Empirically, we found layer \emph{one} to work best for all the language models introduced above. Furthermore, RoBERTa performs \emph{worse} out of the three when comparing the derived static embeddings on this task. In comparison, CUVA performs significantly better, i.e. +4.1\%, +4.5\% and +11.8\% improvement on the \emph{average} of Macro, Micro, and Pair F1 values, when compared against ERNIE+HAC, BERT+HAC and RoBERTa+HAC respectively.

\subsection{Structural Ablations}
\label{structural_ablation}

Table \ref{NellAblations} illustrates the ablation experiments performed on the Entity Canonicalization task for the \textsc{CanonicNell} dataset. The \emph{first} row corresponds to CUVA model used to obtain state-of-the-art results, as reported in Table \ref{NellEntities}. Removing the \emph{hidden} layer out of CUVA's encoder and decoder network, yields the results in the \emph{second} row of the table. The \emph{final} row reports the performance of CUVA without the KGE Module. 

From the results, it is clear that adding an \emph{hidden} layer to the VAE certainly improves CUVA's performance. Moreover, we find that removing the KGE Module drops the Pair F1 value \emph{significantly} by $8.4\%$, with a statistically \emph{insignificant} increase in the Macro and Micro F1 values. This \emph{drop} in Pair F1 is due to a drop in the pairwise precision, i.e., from $0.379$ with KGE to $0.229$ without KGE.

Pairwise precision measures the quality of a set of clusters as the ratio of number of hits to the total possible allowed pairs, wherein a pair of NPs produce a hit if they refer to the same entity. Therefore, using a KGE module causes CUVA to generate a higher hit ratio, and in turn \emph{supports} our hypothesis that a KGE Module helps to better disambiguate entity clusters by considering the context given by the relations, and is therefore necessary.

\subsection{Effectiveness of Joint Learning}
\label{effectiveness_joint_learning}

CUVA models the Canonicalization task via a latent variable generative model and approximates the likelihood of an observed Open KG triple via a variational inference approach. Under this method, the probability of an NP (or RP) belonging to a latent cluster is entangled with both the representations of the observed mentions and the representations of the latent, and consequently, affects the likelihood of an observed triple in a joint manner. This is relevant because it allows gradients to update both the mention embeddings and soft cluster assignments jointly, thereby effectively learning from one another. 

Table \ref{joint_vs_pipeline} empirically demonstrates this relevance, i.e.\ benefits of joint learning over a pipeline approach, while using the same network architecture. In this study, the experiments have been done on the Entity Canonicalization task (\emph{head mentions} only) for the ReVerb45K dataset. In addition to CUVA, we build a second model following a \emph{pipeline} approach, which we refer to as VAE+HAC. This model first uses the same architecture as CUVA for learning mention representations, and in a subsequent independent step, uses a hierarchical agglomerative clustering step to cluster the mentions together. The results indicate that a joint approach outperforms a pipeline-based strategy used by existing state-of-the-art models, such as CESI.

\begin{table}
    \centering 
    \resizebox{.99\columnwidth}{!}{
    \begin{tabular}{l|c|c|c}
    \toprule 
    Approaches & Macro & Micro & Pair \\ \midrule 
    CUVA & 0.775 & 0.826 & 0.363 \\ \midrule 
    Without the Hidden Layer & 0.758 & 0.809 & 0.253 \\ 
    Without the KGE Module & 0.782 & 0.829 & 0.279 \\ 
    \bottomrule
    \end{tabular}}
    \caption{Ablation tests on the Entity Canonicalization task showing Macro, Micro and Pair F1 results for the \textsc{CanonicNell} dataset. All approaches here use the same hyperparameters as in Table \ref{NellEntities}.}
    \label{NellAblations}
\end{table}

\begin{table}
    \centering 
    \begin{tabular}{l|c|c|c}
    \toprule 
    Approaches & Macro F1 & Micro F1 & Pair F1 \\ \midrule 
    CUVA & 0.661 & 0.867 & 0.855 \\
    VAE+HAC & 0.545 & 0.862 & 0.777 \\
    \bottomrule
    \end{tabular}
    \caption{Ablation tests illustrating that our proposed approach for \emph{joint learning} of mention representations and cluster assignments performs better than a \emph{pipeline} approach. The evaluations have been done on the Entity Canonicalization task (\emph{head mentions only}) for the ReVerb45K dataset.}
    \label{joint_vs_pipeline}
\end{table}

\section{Conclusion} \label{conclusion}
In this paper, we introduced CUVA, a novel neural architecture to canonicalize Noun Phrases and Relation Phrases within an Open KG. We argued that CUVA learns unique mention embeddings and cluster assignments in a \emph{joint} fashion, compared to a pipeline strategy followed by the current state of the art methods. Moreover, we also introduced \textsc{CanonicNell}, a new dataset for Entity Canonicalization. An evaluation over four benchmarks demonstrates the effectiveness of CUVA over state of the art baselines. 

\bibliography{anthology,custom}
\bibliographystyle{acl_natbib}

\clearpage
\appendix

\section{Side Information} \label{appendix:SI}
Following CESI \cite{DBLP:conf/www/VashishthJT18_apx}, we use the following five sources of side information, which are described as follows:
\begin{itemize}
    \item \textbf{Entity Linking}: Given unstructured text, from which the triple was extracted, we use Stanford CoreNLP entity linker \cite{DBLP:conf/lrec/SpitkovskyC12} to map Noun Phrases (NPs) to Wikipedia Entities. If two NPs are linked to the same Wikipedia entity, we assume them to be equivalent as per this information. 
    \item \textbf{PPDB Information}: We follow the same strategy as \cite{DBLP:conf/www/VashishthJT18_apx} and modify the PPDB 2.0 \cite{pavlick-etal-2015-ppdb} collection into a set of clusters. If two NPs (or RPs) belong to the same cluster, then they are treated as equivalent.
    \item \textbf{IDF Token Overlap}: In \cite{DBLP:conf/cikm/GalarragaHMS14_apx}, IDF Token Overlap was found to be the most effective feature for canonicalization. For example, it is very likely that \emph{William Shakespeare} and \emph{Shakespeare} refer to the same entity, or in other words, Noun Phrases (NPs) or Relation Phrases (RPs) sharing infrequent terms are more likely to refer to the same entity (or relation). An overlap score for every NP (or RP) pair is calculated as per the formula provided in \cite{DBLP:conf/www/VashishthJT18_apx}, and we keep only those pairs with scores beyond a particular \emph{threshold}.  
    \item \textbf{Morph Normalization}: We use multiple morphological normalization operations, as used in \cite{fader-etal-2011-identifying-apx} for finding out equivalent NPs. 
\end{itemize}

We use the following strategy to calculate the \emph{plausibility scores} for the mention pairs generated by each of the \emph{five} aforementioned sources of Side Information. Mention pairs identified by IDF Token Overlap follow the same \emph{scoring} strategy as mentioned before, whereas mention pairs identified by WordNet (with Word-sense disambiguation) and Morphological normalizations get a score of \emph{one}. 

The remaining sources, i.e. Entity Linking and PPDB, tend to group the NP and RP mentions into clusters. Being empirical in nature, these approaches are \emph{likely} to introduce errors in their results, for e.g.\ due to incorrect disambiguation, and can cause some of the generated clusters to overlap. 

Working with such a set of potentially overlapping clusters, we
make an \emph{observation} that, if a particular \emph{mention} belongs to more than one cluster, then it is likely to be ambiguous, and therefore should have a low equivalence score with other members of the same cluster. Therefore, we score two mentions $p$ and $q$ belonging to the same cluster $\mathcal{C}$ as,
\begin{equation*}
    S_{\mathcal{C}}(p,q) = \frac{1}{|\mathcal{C}|^2}e^{2-(\eta(p)+\eta(q))}
\end{equation*}
where $e$ denotes the exponential function, $\eta(x)$ denotes the number of clusters containing $x$, and $|\mathcal{C}|$ denotes the cluster size. The scaling factor of $1/|\mathcal{C}|^2$ favors clusters of smaller size, since for the \textsc{Canonicalization} task, ideal cluster sizes are expected to be small.

\begin{table*}
    \centering
    \begin{tabular}{c|c|c|c|c|c|c}
    \toprule
         & \multicolumn{3}{c|}{With Normalization} & \multicolumn{3}{c}{Without Normalization} \\ \hline
    Validation fold & Macro & Micro & Pair & Macro & Micro & Pair \\ 
         \midrule
    Base & \bf{1.0} & \bf{1.0} & \bf{1.0} & \bf{1.0} & \bf{1.0} & \bf{1.0} \\
    Ambiguous & 0.811 & 0.966 & 0.964 & \bf{0.823} & \bf{0.976} & \bf{0.966} \\
    ReVerb45K & \bf{0.728} & \bf{0.906} & \bf{0.953} & 0.721 & 0.901 & 0.951 \\ 
    \bottomrule
    \end{tabular}
    \caption{Comparison of \emph{two} initialization strategies when used for initializing mixture of gaussians for running CUVA on the Entity Canonicalization task for \emph{head entity mentions} on the \emph{Validation fold} of the benchmark datasets. The \emph{With Normalization} strategy builds GloVe embeddings for multi-token NPs by averaging unit L2 normalized GloVe vectors for individual tokens, whereas the \emph{Without Normalization} strategy simply averages GloVe vectors for individual tokens to build the embeddings for multi-token NPs. See Section \ref{appendix:initializations} for more details.}
    \label{tab:decide_to_normalize}
\end{table*}

\section{Training Strategy} \label{appendix:trainingDetails}
In this section, we describe our strategy for training the CUVA model. Let $\mathcal{E}, \mathcal{R}$ denote the entity and relation vocabulary for an Open KG. Unless otherwise specified, all \emph{trainable} CUVA parameters are \emph{randomly} initialized. We train the model in \emph{three} stages, as follows:

\subsection{Initializing Mixture of Gaussians} \label{appendix:initializations}

We use the \emph{pretrained} 100-dimensional GloVe vectors \cite{pennington-etal-2014-glove} for embedding matrices $\mathcal{E}_{g}$ and $\mathcal{R}_{g}$ corresponding to the vocabulary $\mathcal{E}$ and  $\mathcal{R}$ respectively. 

The embeddings for multi-token phrases are calculated by averaging GloVe vectors for each token. This step can be done in one of two ways, \begin {enumerate*} [label=\itshape\alph*\upshape)] \item Normalize individual GloVe token vectors and then average them, or \item Average individual GloVe token vectors without Normalizing. \end{enumerate*} In the absence of any other information, we evaluate CUVA on the validation fold of each of the benchmark datasets, as shown in Table \ref{tab:decide_to_normalize}. For each dataset, we mark the embedding initialization strategy that yields the best performance, and then use it to evaluate our model on the test fold of the corresponding benchmark datasets (as illustrated in the main paper).  

Based on the results from Table \ref{tab:decide_to_normalize}, we use the \emph{Without Normalization} strategy for Ambiguous dataset, whereas for ReVerb45K, we use the \emph{With Normalization} strategy. For the Base dataset, both strategies yield the same results and therefore we randomly choose the \emph{With Normalization} strategy and use it while evaluating on the test fold. Furthermore, we choose the \emph{With Normalization} strategy for the \textsc{CanonicNell} dataset as well. 


For the \textsc{Canonicalization} task, the cluster sizes will be likely small, and in turn, we get a large number of clusters. The average-case time complexity per iteration of $k$-Means using Lloyd's algorithm \cite{DBLP:journals/tit/Lloyd82} is $O(nk)$, where $n$ is the number of samples. However, for our case, as $k$ is comparable to $n$, the average time complexity becomes $O(n^2)$ similar to the Hierarchical Agglomerative Clustering (HAC) method with complete linkage criterion \cite{DBLP:journals/cj/Defays77}. Though both methods have the same time complexity, we use HAC as our clustering method as we observe that it gives a better performance empirically. We cover the empirical comparison between both methods of initialization, i.e. HAC and KMeans in Section \ref{appendix:OtherInitializations} of this Appendix. 

We run HAC separately over $\mathcal{E}_{g}$ for NPs, and $\mathcal{R}_{g}$ over RPs. We use two different thresholds $\theta_E$ for entities, and $\theta_R$ for relations to convert the output dendrograms from HAC into \emph{flat} clusters. Using these clusters, we compute within-cluster means and variances to initialize the \emph{means} and the \emph{variances} of the Gaussians for both E-VAE and R-VAE respectively. Note that, the choice of $\theta_E$ and $\theta_R$ sets the values for the number of mixtures $K_{E}$ and $K_{R}$ used in the next stage.

\subsection{Two-step training procedure} \label{appendix:TwoStepTraining}

We train CUVA in \emph{two} independent steps. Our training strategy is similar to \cite{DBLP:conf/icml/MiaoYB16} where they train the encoder and decoder of the VAE alternatively rather than simultaneously. In the first step, we train the encoder in both E-VAE and R-VAE while keeping the decoder fixed. Then, in the second step, we keep the encoder fixed and only train the decoder. 

\paragraph{Encoder training:} We train the \emph{Encoder} for both E-VAE and R-VAE by using the labels generated via the HAC algorithm (during initialization of the mixture of gaussians) as a source of weak supervision. Specifically, for a given triple $(h,r,t)$, we compute: 
\begin{itemize}
    \item Negative log likelihood (NLL) loss $\mathcal{L}_h$ calculated using the predicted cluster assignment probability vector for $h$ and the cluster label for $h$.
    \item NLL values $\mathcal{L}_r, \mathcal{L}_t$ for $r,t$ computed in a similar manner.
    \item L1 Regularizer values using the \emph{Encoder} parameters for E-VAE and R-VAE, denoted by $\mathcal{L}_{\text{REG1}}$.
    \item Side Information Loss $\mathcal{L}_{\text{SI}}$ applicable between any two equivalent NPs (or RPs). See Figure \ref{architecture}.
\end{itemize}
The \emph{overall} loss function for the \emph{first} step is therefore, 
\begin{equation*} \label{train:stage1}
    \mathcal{J} = \sum_{(h,r,t) \in \mathcal{T}} \mathcal{L}_h + \mathcal{L}_r + \mathcal{L}_t + \lambda \mathcal{L}_{\text{REG1}} + \mathcal{L}_{\text{SI}}
\end{equation*}

We train the \emph{Encoder} for a maximum of $T_e$ epochs, and then proceed to the \emph{second} step.

Using labels generated by the HAC algorithm as a source of weak supervision introduces noise and sets an upper limit to how much CUVA can learn. However, we also use side information during the Encoder training procedure, which helps CUVA fix the errors introduced by HAC, thus resulting in an \emph{improved} performance. This behavior is empirically demonstrated by comparing GloVe+HAC and CUVA approaches on the ReVerb45K dataset in the main paper.

\paragraph{Decoder training:} In this step, we train the \emph{decoder} only, and keep the \emph{encoder} fixed. The cluster parameters and the embedding lookup table are also updated. The decoder is trained by minimizing the following \emph{loss} values:
\begin{itemize}
    \item The evidence lower bound (ELBO) loss $\mathcal{L}^E_{\text{ELBO}}$ for E-VAE and $\mathcal{L}^R_{\text{ELBO}}$ for R-VAE respectively, with the \emph{decoder} being a multivariate Gaussian with a diagonal covariance structure. The ELBO loss breaks into two parts namely, the Reconstruction Loss, and the KL divergence between the variational posterior and the prior. The expressions for ELBO loss are based on \cite{DBLP:conf/ijcai/JiangZTTZ17_apx}.
    
    \item The KGE Module loss $\mathcal{L}_{\text{KGE}}$ and the Side Information Loss $\mathcal{L}_{\text{SI}}$ (Refer to Figure \ref{architecture}).
    
    \item L1 Regularizer loss values ($\mathcal{L}_{\text{REG2}}$) using the \emph{Decoder} parameters for E-VAE and R-VAE.
\end{itemize}

The \emph{combined} loss function for the \emph{second} step is: 
\begin{equation*}
    \mathcal{J} = \mathcal{L}^E_{\text{ELBO}} + \mathcal{L}^R_{\text{ELBO}} + \mathcal{L}_{\text{KGE}} +\mathcal{L}_{\text{SI}} +  \lambda \mathcal{L}_{\text{REG2}}
\end{equation*}

where $\lambda$ corresponds to the \emph{weight} value for the regularizer, a hyper-parameter set to $0.001$. The decoder is trained for a maximum of $T_d$ epochs.

The motivation behind using a two-step training strategy for the VAEs is to prevent the decoder from ignoring latent representations $z$ and learning directly from the input data \cite{bowman-etal-2016-generating}. Once the encoder has been trained in the first step, we keep the encoder weights fixed for the second step. This forces the decoder to learn only from the \emph{latent} representations, and not from the input data. Note that the KGE loss $\mathcal{L}_{\text{KGE}}$ is not used in Step one, since it causes the model to diverge in practice.

\section{Hyperparameters} \label{appendix:Hyperparameters}

In this section, we discuss the grid search for hyperparameters and present the final hyperparameters used.

\subsection{Grid Search Details} \label{appendix:GridSearchDetails}
The search space used to obtain the best performing hyper-parameters for our experiments is described as follows: We calculate the \emph{threshold} cutoff for HAC based initializations using the validation fold via a two-step approach. In the \emph{first} step, we use a search space of  $[0.2, 1.0)$ in steps of $0.1$. In the \emph{final} step, we take the best cutoff value $c$ from the previous step and construct a new search space $[c-0.1, c+0.1]$ with a step size of $0.01$. Finally, we take the best performing threshold cutoff value from the previous step and use it to evaluate CUVA models on the test set. 

For choosing the \emph{threshold} cutoff for the IDF Token Overlap strategy in regards to Entity Side Information, we used a search space of $[0.2, 0.8]$ in increments of $0.1$ for all the datasets. In comparison, we chose $0.9$ as a cutoff for the IDF Token Overlap strategy in regards to Relation Side Information (wherever applicable), without any search as it already produced a decent number of relation pairs and manual inspection of a sample indicated good quality. Finally, as to the choice of the latent space dimensions for the VAE, we employed a grid search over $\{50, 100, 200\}$ dimensions.

\subsection{Final Hyperparameters used}

We use the following hyperparameter values in our experiments.

\paragraph{Common hyperparams.}
The fully connected layers in the Encoder section of the VAEs have embedding dimensions of 768, 384, and 100, whereas the Decoder sections have the same dimensions, but in reverse order. Both Encoder and Decoder use \emph{tanh} nonlinearities. A learning rate of 1e-3 and 1e-4 together with Adam optimizer \cite{DBLP:journals/corr/KingmaB14} is used in steps one and two during our proposed two-step training procedure. L1 regularization with a regularizer weight of 1e-3 is used. A batch size of 50 is used for training, whereas for evaluation, we use a batch size of 5. Moreover, we use 20 random negative samples per positive sample, while calculating the loss function pertaining to the HolE algorithm. The GloVe vectors used for initializing the Gaussian Mixture models are obtained from \url{http://nlp.stanford.edu/data/GloVe.6B.zip}.

For Base, Ambiguous and ReVerb45K datasets, we use a threshold of 0.4 for entities and 0.9 for relations regarding the IDF Token Overlap strategy for scoring Side Information pairs, i.e. pairs whose scores are less than these cutoff values, are discarded. For \textsc{CanonicNell} we employ a threshold of 0.5 concerning the IDF Token Overlap strategy for scoring Entity Side Information pairs. Furthermore, the relations within \textsc{CanonicNell} are unique, therefore they are treated as singleton clusters for the experiments.

\paragraph{Dataset specific hyperparameters.} The dataset specific hyperparameters are illustrated in Table \ref{tab:dataset_specific_hyperparams}. The first six rows corresponds to hyperparameters related to our proposed CUVA model, whereas the final row pertains to the Seed values (for reproducibility purposes) used for evaluating CUVA models on the test fold of the benchmark datasets.

\begin{table}
    \centering
    \resizebox{.99\columnwidth}{!}{
    \begin{tabular}{c|c|c|c|c}
    \toprule
    Params & Base & Ambiguous & ReVerb45K & \textsc{CanonicNell} \\ \midrule
    $\theta_{E}$ & 0.53 & 0.3 & 0.4 & 0.21 \\ 
    $\theta_{R}$ & 0.43 & 0.5 & 0.37 & N/A \\ 
    $K_{E}$ & 1021 & 5013 & 12965 & 6625 \\ 
    $K_{R}$ & 102 & 625 & 1076 & N/A \\ 
    $T_{e}$ & 50 & 50 & 50 & 50 \\ 
    $T_{d}$ & 300 & 300 & 300 & 100 \\ 
    Seed & 42 & 57 & 55 & 10 \\ 
    \bottomrule
    \end{tabular}}
    \caption{Final dataset specific hyperparameters used for training and evaluating CUVA models on the test fold of these benchmark datasets. Here, $\theta_{E}$ and $\theta_{R}$ denote the threshold cutoff used during HAC based initializations. Setting the values of $\theta_{E}$ and $\theta_{R}$ sets the values for the number of mixtures $K_{E}$ and $K_{R}$ used in the E-VAE and R-VAE respectively. See Section \ref{appendix:initializations} for a detailed description on the notations used.}
    \label{tab:dataset_specific_hyperparams}
\end{table}

Moreover, all experiments are implemented in PyTorch v1.4.0 using a single Intel x86 CPU and one NVIDIA v100 GPU, with a max of 16GB RAM.

\section{Other Ablation Experiments} \label{appendix:OtherInitializations}

In this section, we describe additional experiments to analyze the performance of our proposed CUVA model.

Table \ref{OtherInitializationsTable} illustrates the performance of CUVA while varying the strategies on the choice of initializations for the Gaussian Mixture model and Knowledge Graph Embedding. While our proposed instantiation of CUVA, i.e. Row \emph{two}, uses HAC clustered GloVe vectors for initializations and HolE for Knowledge Graph Embedding, it is worthwhile to note that all the other combinations also do outperform CESI, which is the current state of the art model. 

\begin{table}[ht]
    \centering 
    \resizebox{.999\columnwidth}{!}{
    \begin{tabular}{c|c|c|c}
    \toprule 
    & Macro F1 & Micro F1 & Pair F1 \\ \midrule 
    CESI & 0.627 & 0.844 & 0.819 \\
    CUVA & 0.661 & \bf{0.867} & \bf{0.855} \\ \midrule  
    Glove+TransE+HAC & \bf{0.662} & 0.862 & 0.837 \\ 
    Glove+HolE+KMeans & 0.633 & 0.855 & 0.826 \\ 
    FastText+HolE+HAC & 0.651 & 0.858 & 0.823 \\
    \bottomrule
    \end{tabular}}
    \caption{Results on the performance of CUVA while using other initialization strategies. The results are reported on the \emph{Entity Canonicalization} task for \emph{head entities only} on ReVerb45k.}
    \label{OtherInitializationsTable}
\end{table}

\begin{figure}[h]
  \centering
  \includegraphics[width=\columnwidth]{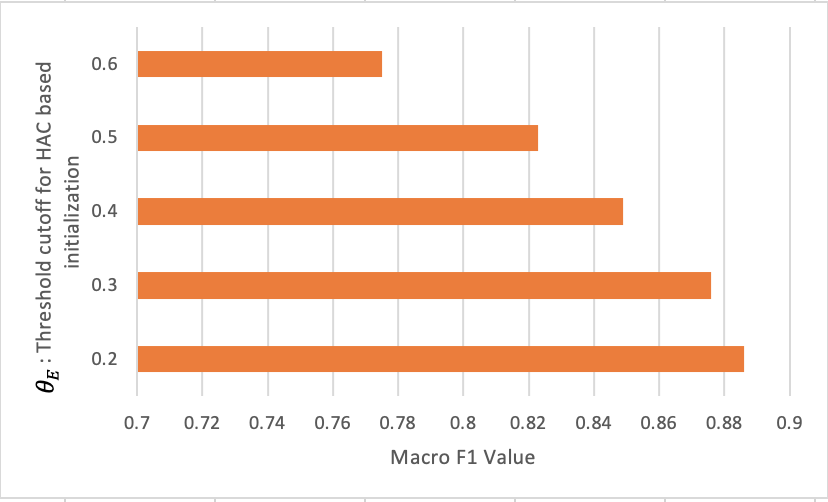}
  \caption{Comparison of Macro F1 results for the Entity Canonicalization task on \emph{all entity} mentions for the Ambiguous dataset as a function $\theta_{E}$, where $\theta_{E}$ denotes the threshold cutoff used during HAC based initialization and in turn sets the value for the number of entity clusters used in CUVA. All other hyperparams remain identical to the values denoted in Table \ref{tab:dataset_specific_hyperparams}. The list of Macro F1 values reported here have a standard deviation of $4.6\%$.} 
  \label{appendix:hyperparam_sensitivity}
\end{figure}

Figure \ref{appendix:hyperparam_sensitivity} illustrates the comparison of Macro F1 results for the Entity Canonicalization task on \emph{all entity} mentions for the \emph{test fold} of the Ambiguous dataset as a function of $\theta_{E}$. Here, $\theta_{E}$ denotes the threshold cutoff used during HAC based initializations, which in turn sets the value for the number of mixtures $K_{E}$ in CUVA. We donot highlight the Micro or Pair F1 values, since the relative change in those values while varying $\theta_{E}$ was minimal.

Its interesting to note that setting $\theta_{E}=0.2$ yields a better Macro F1 value (by $1\%$) on the \emph{test} fold, even though $\theta_{E}=0.3$ had the best performance on the \emph{validation} fold.

Furthermore, from Section \ref{appendix:SI} (of the Appendix), we note that CUVA uses several sources to generate additional side information, and utilizes it while training. Specifically, CUVA uses two \emph{external} resources, i.e.\  an off the shelf Stanford CoreNLP entity linker (EL) \cite{DBLP:conf/lrec/SpitkovskyC12} and a lexical resource called PPDB 2.0 \cite{pavlick-etal-2015-ppdb}, which is a collection of equivalent paraphrases. Table \ref{SI_sources_ablations} illustrates the performance of CUVA when each of these \emph{external} resources is ablated one at a time. 

From these results, it is clear that the entity linker (EL) has a bigger impact on the results as opposed to PPDB 2.0 resource. This is because, being a statistical model, the Stanford CoreNLP entity linker is much more likely to capture lexical variations within mentions of the same entity, as opposed to the PPDB 2.0 lexical resource.

\begin{table}
    \centering 
    \resizebox{.99\columnwidth}{!}{
    \begin{tabular}{l|c|c|c}
    \toprule 
    Approaches & Macro & Micro & Pair \\ \midrule 
    CUVA & 0.661 & 0.867 & 0.855 \\ \midrule
    Without PPDB2 & 0.668 & 0.867 & 0.841 \\
    Without EL & 0.595 & 0.845 & 0.846 \\ 
    Without EL and PPDB2 & 0.592 & 0.844 & 0.827 \\ 
    \bottomrule
    \end{tabular}}
    \caption{Ablation tests demonstrating the effects of using external \emph{resources} on the Entity Canonicalization task (\emph{head mentions only}) for the ReVerb45K dataset.}
    \label{SI_sources_ablations}
\end{table}

\end{document}